\documentclass{article}

     \PassOptionsToPackage{numbers, compress}{natbib}

     \usepackage[preprint]{neurips_2022}

\usepackage[utf8]{inputenc} 
\usepackage[T1]{fontenc}    
\usepackage{hyperref}       
\usepackage{url}            
\usepackage{booktabs}       
\usepackage{amsfonts}       
\usepackage{nicefrac}       
\usepackage{microtype}      
\usepackage{xcolor}         

\usepackage{amsmath}
\usepackage{amssymb}
\usepackage{mathtools}
\usepackage{amsthm}

\usepackage{algorithmic}

\usepackage{tikz}
\usetikzlibrary{automata, positioning, arrows}
\tikzset{
    ->, 
    >=stealth', 
    node distance=2cm, 
    initial text=$ $, 
}
\usepackage{physics}
\usepackage{paralist}
\usepackage{multirow}
\usepackage{multicol}
\usepackage{caption}
\usepackage{subcaption}

\usepackage{blkarray}

\title{Global Normalization for Streaming Speech Recognition in a Modular Framework}

\author{
  Ehsan Variani, Ke Wu, Michael Riley, David Rybach, Matt Shannon, Cyril Allauzen \\
  \\
  Google Research \\
  \\
  \texttt{\{variani,wuke,riley,rybach,mattshannon,allauzen\}}@google.com
}

\begin{document}

\maketitle

\begin{abstract}
  We introduce the Globally Normalized Autoregressive Transducer (GNAT) for addressing the label bias problem in streaming speech recognition.
  Our solution admits a tractable exact computation of the denominator for the sequence-level normalization.
  Through theoretical and empirical results, we demonstrate that by switching to a globally normalized model, the word error rate gap between streaming and non-streaming speech-recognition models can be greatly reduced (by more than 50\% on the Librispeech dataset).
  This model is developed in a modular framework which encompasses all the common neural speech recognition models.
  The modularity of this framework enables controlled comparison of modelling choices and creation of new models.
\end{abstract}
\section{Introduction}
\label{sec:introduction}

Deep neural network models have been tremendously successful in the field of automatic
speech recognition (ASR). Several different models have been proposed over the years:
cross-entropy (CE) models with a deep feed-forward architecture \cite{hinton2012deep},
connectionist temporal classification (CTC) models \cite{graves2006connectionist} with recurrent
architectures such as long short-term memory (LSTM) \cite{hochreiter1997long}, and more recently
sequence-to-sequence (Seq2Seq) models like listen, attend and spell (LAS) \cite{chan2015listen},
recurrent neural network transducer (RNN-T) \cite{graves2012sequence}, and
hybrid autoregressive transducer (HAT) \cite{variani2020hybrid}.
When configured in non-streaming mode, these neural ASR models have reached state-of-the-art
word error rate (WER) on many tasks. However, the WER significantly drops when they are operating
in streaming mode. In this paper, we argue that one main cause of such WER gap is that
all the existing models are constrained to be locally normalized
which makes them susceptible to \emph{label bias} problem \cite{smith2007weighted, andor2016globally, lafferty2001conditional, bottou1991approche}.
To address this problem, we introduce new category of globally normalized models called
\emph{Globally Normalized Autoregressive Transducer} (GNAT). Our contributions are:
\begin{compactenum}[(1)]
    \item {\bf Addressing the label bias problem in streaming ASR through global normalization} that significantly closes more than 50\% of the WER gap between streaming and non-streaming ASR.
    \item {\bf Efficient, accelerator-friendly algorithms for the exact computation of the global normalization} under the finite context assumption.\footnote{The implementation is included in the supplementary material of this paper.}
    \item {\bf A modular framework for neural ASR} which encompasses all the common models (CE, CTC, LAS, RNN-T, HAT), allowing creation of new ones, and extension to their globally normalized counterparts.
\end{compactenum}

\section{Streaming Speech Recognition}

For an \emph{input feature sequence} $\vb{x} = x_1 \dots x_T$, usually represented as a sequence of real valued feature vectors (such as log mel), and a finite output alphabet $\Sigma$, we wish to predict the corresponding \emph{output label sequence} $\vb{y} = y_1 \dots y_U,\ y_i \in \Sigma$.
We call each element $x_i$ in $\vb{x}$ a \emph{frame}, and each element $y_i$ in $\vb{y}$ an \emph{output label}.
Common ASR models do not directly predict $\vb{y}$, but rather a \emph{alignment label sequence} $\vb{z} = z_1 \ldots z_V, z_i \in \Sigma \cup \Delta$.
$\Delta$ is a finite alphabet of control labels, such as the blank label in CTC or RNN-T, or the end-of-sequence label in LAS.
There is a deterministic mapping $y(\vb{z}): (\Sigma \cup \Delta)^* \rightarrow \Sigma^*$ for obtaining $\vb{y}$ from $\vb{z}$ (e.g. in RNN-T, we simply remove all the blank labels from $\vb{z}$).
An ASR model can then be broken down in two tasks:
\begin{inparaenum}[(1)]
  \item assigning a score $\prod_i \omega(z_i | \vb{x}, \vb{z}_{<i})$ to each alignment sequence $\vb{z}$, where $\omega(z_i | \vb{x}, \vb{z}_{<i}) \geq 0$ is the \emph{alignment score} of predicting a single alignment label;
  \item finding (usually approximately) $\arg\max_{\vb{y}} \sum_{z|y(\vb{z}) = \vb{y}} \prod_i \omega(z_i | \vb{x}, \vb{z}_{<i})$.
\end{inparaenum}

A non-streaming ASR model's alignment score $\omega(z_i | \vb{x}, \vb{z}_{<i})$ has access to the entire $\vb{x}$ for any $i$.
In contrast, a streaming model's alignment score takes the form $\omega(z_i | \vb{x}_{<t(i)}, \vb{z}_{<i})$: it only has access a prefix $\vb{x}_{<t(i)}$ of the input feature sequence, where $t(i)$ is the frame to which $z_i$ is \emph{aligned}.
Streaming models can be seen as special case of non-streaming models with respect to the alignment scores.

All the common neural ASR models use a locally normalized alignment score which satisfies the constraint $\sum_{z \in \Sigma \cup \Delta} \omega(z_i = z| \vb{x}, \vb{z}_{<i}) = 1$.
This is achieved by applying the softmax function to the last layer activations.
The local normalization constraint makes $\omega(z_i | \vb{x}, \vb{z}_{<i})$ easily interpretable as a conditional probability distribution $P_{\omega}(z_i | \vb{x}, \vb{z}_{<i})$, and thus $\prod_i \omega(z_i | \vb{x}, \vb{z}_{<i})$ easily interpretable as $P_{\omega}(\vb{z} | \vb{x})$.
The modeling parameters are optimized by minimizing the negative log-conditional-likelihood loss $E_{P(\vb{x},\vb{y})}[-\log P_{\omega}(\vb{y}|\vb{x})] = E_{P(\vb{x},\vb{y})}[-\log\sum_{\vb{z}|y(\vb{z}) =\vb{y}} P_{\omega}(\vb{z} | \vb{x})]$.

\subsection{Label Bias in Streaming ASR}

For a non-streaming, locally normalized model, the negative log-conditional-likelihood loss is minimized by setting $\omega(z_i|\vb{x}, \vb{z}_{<i})$ to the true conditional probability $P(z_i|\vb{x}, \vb{z}_{<i})$, leading to $\prod_i \omega(z_i|\vb{x}, \vb{z}_{<i})$ being equal to the true posterior probability $P(\vb{z}|\vb{x})$.

For a streaming model, $\vb{x}$ is replaced by $\vb{x}_{<t(i)}$ in the alignment score (e.g. by using a unidirectional encoder).
Here the negative log-conditional-likelihood loss is minimized by setting $\omega(z_i|\vb{x}_{<t(i)}, \vb{z}_{<i})$ to $P(z_i|\vb{x}_{<t(i)}, \vb{z}_{<i})$.
As a result, the product $\prod_i \omega(z_i|\vb{x}_{<t(i)}, \vb{z}_{<i}) = \prod_i P(z_i|\vb{x}_{<t(i)}, \vb{z}_{<i})$ is in general not equal to $P(\vb{y} | \vb{x})$ anymore.
In other words, using a streaming locally normalized model means that the estimated alignment sequence posterior is the product of some locally normalized alignment scores which depend only on partial input ${\vb{x}}_{<t(i)}$, and as a result can no longer accurately represent the true conditional distribution.
This will bias the model towards predictions with low-entropy estimated posterior probabilities at each decoding step. This degrades the model ability to revise previous decisions, a phenomenon called label bias \cite{lafferty2001conditional, andor2016globally}.

\subsection{Global Normalization}

Traditionally, \emph{globally normalized models} such as conditional random fields \cite{lafferty2001conditional} are used to address the label bias problem.
This paper seeks to apply global normalization to modern neural architectures that are more similar to CTC, RNN-T, or LAS, rather than traditional linear models with the purpose of addressing label bias problem for streaming ASR models.

A globally normalized model does not constrain the alignment score $\omega(z_i | \vb{x}, \vb{z}_{<i})$ to be locally normalized; it only requires it to be any non-negative score, as long as the \emph{denominator} $Z(\vb{x}) = \sum_{\vb{z}} \prod_i \omega(z_i | \vb{x}, \vb{z}_{<i})$ is finite.
The finite denominator allows us to interpret $\frac{\prod_i \omega(z_i | \vb{x}, \vb{z}_{<i})}{Z(\vb{x})}$ as a conditional probability distribution $P_{\omega}(\vb{z} | \vb{x})$.
It is worth noting that any locally normalized model is trivially a globally normalized because $Z(\vb{x}) = 1$ in this case.
Minimum negative log-conditional-likelihood training can be more expensive for globally normalized models due to the need to compute $Z(\vb{x})$ and the corresponding gradients.
However with our proposed modular framework, globally normalized model training can be made practical with careful modelling choices on modern hardware.

For non-streaming settings, \cite{smith2007weighted} shows that locally and globally normalized models express the same class of conditional distributions $P(\vb{z}|\vb{x})$.
Based on this observation, we argue that under non-streaming settings, with adequately powerful neural architectures, maximum log-conditional-likelihood training should yield behaviorly similar locally or globally normalized models, and thus similar WERs in testing.
Results from \cite{hannun2020differentiable} and our own experiments in Section \ref{sec:experiments} validate this.

\section{A Modular Framework for Neural ASR} \label{sec:gnat-model}

In this section, we introduce a modular framework for neural ASR, using the weighted finite state automaton (WFSA) formalism to calculate the conditional probabilities via alignment scores. The modular framework clearly expresses the modelling choices enabling practical globlally normalized model training and inference.
We use the WFSA formalism as the language for describing our framework because of its succinctness and precision, even though our algorithms cannot be directly implemented using existing toolkits such as OpenFst \cite{allauzen2007openfst} or Kaldi \cite{povey2011kaldi}.

\subsection{Preliminaries} \label{subsec:preliminaries}

We begin with an introduction to the relevant concepts and notations.

A \emph{semiring} $(\mathbb{K}, \oplus, \otimes, \bar{0}, \bar{1})$ consists of a set
$\mathbb{K}$ together with an associative and commutative operation $\oplus$
and an associative operation $\otimes$, with respective identities $\bar{0}$ and
$\bar{1}$, such that $\otimes$ distributes over $\oplus$, and
$\bar{0} \otimes x = x \otimes \bar{0} = \bar{0}$.
The \emph{real} semiring $(\mathbb{R}_+, +, \times, 0, 1)$ is used when
the weights represent probabilities. The \emph{log} semiring $(\mathbb{R} \cup
\{\infty\} , \oplus_{\operatorname{log}}, +, \infty, 0)$, isomorphic to the
real semiring via the negative-log mapping, is often used in practice
for numerical stability.\footnote{$a \oplus_{\operatorname{log}} b = -\log(e^{-a}+e^{-b})$}
The \emph{tropical} semiring $(\mathbb{R} \cup
\{\infty\} , \operatorname{min}, +, \infty, 0)$
is often used in shortest-path applications. 

A \emph{weighted finite-state automaton} (WFSA) $A = (\Sigma, Q, i,
F, \rho, E)$ over a semiring $\mathbb{K}$ is specified by a finite 
alphabet $\Sigma$, a finite set of
states $Q$, an initial state $i \in Q$, a set of final states $F \subseteq Q$,
a final state weight assignment $\rho : F \rightarrow \mathbb{K}$, and a finite
set of transitions $E \subseteq Q \times (\Sigma \cup \{\epsilon\})
\times \mathbb{K} \times Q$ ($\epsilon$ denotes the empty label sequence).
Given a
transition $e \in E$, $p[e]$ denotes its origin or previous state, $n[e]$ its
destination or next state, $o[e]$ its label, and
$\omega[e]$ its weight. A \emph{path} $\pi = e_1 \dots e_k$ is a sequence of
consecutive transitions $e_i \in E$: $n[e_{i-1}] = p[e_i],\, i = 2, \dots k$.
The functions $n$, $p$, and $\omega$ on transitions can be extended to paths by
setting: $n[\pi] = n[e_k]$ and $p[\pi] = p[e_1]$ and by defining the weight of
a path as the $\otimes$-product of the weights of its constituent transitions:
$\omega[\pi] = \omega[e_1] \otimes \dots \otimes \omega[e_k])$.
An \emph{unweighted finite-state automaton} (FSA) $A = (\Sigma, Q, i, F, E)$ is simply a WFSA whose transitions and final states are all weighted by $\bar{1}$.

$\Pi(Q_1, Q_2)$ is the set of all paths from a subset $Q_1 \subseteq Q$ to a
subset $Q_2\subseteq Q$. $\Pi(Q_1, \vb{y}, Q_2)$ is the subset of all paths of
$\Pi(Q_1, Q_2)$ with label sequence $\vb{y} = y_1 \dots y_U$, $y_i \in \Sigma$.
A path in $\Pi(\{i\}, F)$ is said to be accepting or
\emph{successful}.
The weight associated by $A$ to any label sequence $\vb{y}$ is
given by $A(\vb{y}) = \bigoplus_{\pi \in \Pi(\{i\}, \vb{y}, F)} \omega[\pi] \otimes \rho(n[\pi])$.
The \emph{weight} of $A$ is the $\oplus$-sum of weights of all accepting paths $W(A) = \bigoplus_{\pi \in \Pi(\{i\}, F)} \omega[\pi] \otimes \rho(n[\pi])$.
For a semiring $\mathbb{K}$ where $\otimes$ is also commutative, the \emph{intersection} (or Hadamard product) of two WFSA $A_1$ and $A_2$ is defined as: $(A_1 \cap A_2)(\vb{y}) = A_1(\vb{y}) \otimes A_2(\vb{y})$.
\cite{mohri2009} gives an algorithm to compute the intersection.
We can view $\vb{y}$ as a WFSA that accepts only $\vb{y}$ with weight $\bar{1}$, then $A(\vb{y}) = W(A \cap \vb{y})$.

\subsection{Probabilistic Modeling and Inference on Acyclic Recognition Lattices} \label{subsec:probabilistic-modeling-and-inference}

For any feature sequence $\vb{x}=x_1 \dots x_T$, a model with trainable parameters $\theta$ induces a \emph{recognition lattice} WFSA $A_{\theta,\vb{x}} = (\Sigma, Q_{\theta,\vb{x}}, i_{\theta,\vb{x}}, F_{\theta,\vb{x}}, \rho_{\theta,\vb{x}}, E_{\theta,\vb{x}})$.
For a label sequence $\vb{y} = y_1 \dots y_U$, the recognition lattice $A_{\theta,\vb{x}}(\vb{y})$ under the log semiring can be viewed as the unnormalized negative log conditional probability $P_{\theta}(\vb{y} \mid \vb{x}) = \frac{\exp\left(-W(A_{\theta,\vb{x}}(\vb{y}))\right)}{\exp\left(-W(A_{\theta,\vb{x}})\right)}
        = \frac{\exp\left(-W(A_{\theta,\vb{x}} \cap \vb{y})\right)}{\exp\left(-W(A_{\theta,\vb{x}})\right)}$.
        
The recognition lattice $A_{\theta,\vb{x}}$ is designed to be acyclic,
and therefore the weight of the automata in both the numerator and denominator above can be efficiently computed by visiting the states of the corresponding WFSA in topological order \cite{mohri2002}.
See Appendix~\ref{sec:accelerator_friendly_modeling} for our accelerator friendly version of this algorithm.
We can thus train an ASR model by minimizing the negative log-conditional-likelihood on the training corpus $\mathcal{D}$, and choosing
$\theta^\star = \operatorname{arg\,min}_{\theta} \mathop{\mathbb{E}}_{(\vb{x}, \vb{y}) \in \mathcal{D}} [-\log(P_{\theta}(\vb{y} \mid \vb{x}))] = \operatorname{arg\,min}_{\theta} \mathop{\mathbb{E}}_{(\vb{x}, \vb{y}) \in \mathcal{D}} [ W(A_{\theta,\vb{x}} \cap \vb{y}) - W(A_{\theta,\vb{x}}) ]$.

In general, there can be more than one path in $A_{\theta,\vb{x}}$ that accepts the same $\vb{y}$.
During inference, finding the optimal $\hat{\vb{y}} = \operatorname{arg\,max}_{\vb{y}} P_{\theta}(\vb{y}|\vb{x})$ requires running the potentially expensive WFSA disambiguation algorithm \cite{mohri2015disambiguation} on $A_{\theta,\vb{x}}$.
As a cheaper approximation, we instead look for the shortest path $\hat{\pi}$ in $A_{\theta,\vb{x}}$ under the tropical semiring, and use the corresponding label sequence as the prediction, again using the standard shortest path algorithm for an acyclic WFSA \cite{mohri2002}.

\subsection{Inducing the WFSA} \label{subsec:inducing-the-wfsa}

Our framework decomposes the sequence prediction task in ASR into three components, each playing a specific role in inducing the recognition lattice $A_{\theta,\vb{x}}$.
\begin{compactitem}
    \item The \emph{context dependency} FSA $C = (\Sigma, Q_C, i_C, F_C, E_C)$ is an $\epsilon$-free, unweighted FSA, whose states encode the history of the label sequence produced so far. $C$ is fixed for a given GNAT model, independent of input $\vb{x}$.
    \item The \emph{alignment lattice} FSA $L_T = (\Sigma, Q_T, i_T, F_T, E_T)$ is an acyclic, unweighted FSA, whose states encode the alignment between input frames $\vb{x}$ and output labels $\vb{y}$. $L_T$ depends on only the length $T$ of input $\vb{x}$.
    \item The \emph{weight} function $\omega_{\theta,\vb{x}}: Q_T \times Q_C \times (\Sigma \cup \epsilon) \rightarrow \mathbb{K}$. $\omega_{\theta,\vb{x}}$ is the only component that contains trainable parameters and requires full access to $\vb{x}$. This function defines the transition weights in the recognition lattice $A_{\theta,\vb{x}}$.
\end{compactitem}
We will discuss how one can define these components in detail in the next section.
With $(C, L_T, \omega_{\theta,\vb{x}})$ given, the recognition lattice $A_{\theta,\vb{x}}$ is defined as follows:
\begin{align*}
    Q_{\theta,\vb{x}} & = Q_T \times Q_C \\
    i_{\theta,\vb{x}} & = (i_T, i_C) \\
    F_{\theta,\vb{x}} & = F_T \times F_C \\
    \begin{split}
        E_{A_{\theta,\vb{x}}}  = & \Bigl\{\bigl((q_a, q_c), y, \omega_{\theta,\vb{x}}(q_a, q_c, y), (q_a', q_c')\bigr) \mid \\ & y \in \Sigma,\ (q_a, y, q'_a) \in E_T,\ (q_c, y, q'_c) \in E_C\Bigr\} \\
            &  \cup \Bigl\{\bigl((q_a, q_c), \epsilon, \omega_{\theta,\vb{x}}(q_a, q_c, \epsilon), (q_a', q_c)\bigr) \mid \\
            & (q_a, \epsilon, q'_a) \in E_T,\  q_c \in Q_C\Bigr\}
    \end{split}  \\
    \rho_{A_{\theta,\vb{x}}}(q) & = \bar{1},\ \forall q \in F_{\theta,\vb{x}}
\end{align*}
In other words, the topology (states and unweighted transitions) of the recognition lattice $A_{\theta,\vb{x}}$ is the same as the FSA intersection $L_T \cap C$; and the transition weights are defined using $\omega_{\theta,\vb{x}}$.
The $\epsilon$-freeness of $C$ and the acyclicity of $L_T$ implies that the recognition lattice $A_{\theta,\vb{x}}$ is also acyclic.

\section{Components of a GNAT Model} \label{sec:components-of-a-gnat-model}

In this section, we define \emph{globally normalized autoregressive transducer} (GNAT) through the framework above, by specifying each model component.

\begin{figure}
    \centering
    \begin{subfigure}[b]{0.45\textwidth}
        \centering
        \scalebox{0.75}{%
        \begin{tikzpicture}
            \node[state, initial, accepting] (empty) {$\epsilon$};
            \node[state, accepting, below left of=empty, yshift=-1cm] (a) {$a$};
            \node[state, accepting, below right of=empty, yshift=-1cm] (b) {$b$};
    
            \draw (empty) edge[bend right, above left] node{$a$} (a)
                  (empty) edge[bend left, above right] node{$b$} (b)
                  (a) edge[loop left] node{$a$} (a)
                  (a) edge[bend left, above] node{$b$} (b)
                  (b) edge[bend left, above] node{$a$} (a)
                  (b) edge[loop right] node{$b$} (b);
        \end{tikzpicture}
        }%
        \caption{$n$-gram context-dependency automaton $C_2$ for $\Sigma=\{a,b\}$.}
        \label{fig:n-gram-context}
    \end{subfigure}
    \hfill
    \begin{subfigure}[b]{0.45\textwidth}
        \centering
        \scalebox{0.75}{%
        \begin{tikzpicture}
            \node[state, initial] (q0) {0};
            \node[state, right of=q0] (q1) {1};
            \node[state, right of=q1] (q2) {2};
            \node[state, accepting, right of=q2] (q3) {3};
    
            \draw (q0) edge[bend right, above] node{$\Sigma$} (q1)
                  (q0) edge[bend left, above] node{$\epsilon$} (q1)
                  (q1) edge[bend right, above] node{$\Sigma$} (q2)
                  (q1) edge[bend left, above] node{$\epsilon$} (q2)
                  (q2) edge[bend right, above] node{$\Sigma$} (q3)
                  (q2) edge[bend left, above] node{$\epsilon$} (q3);
        \end{tikzpicture}
        }
        \caption{Frame dependent alignment lattice with $T=3$.}
        \label{fig:frame-dependent}
    \end{subfigure}
    \hfill
    \begin{subfigure}[b]{0.45\textwidth}
        \centering
        \scalebox{0.6}{%
        \begin{tikzpicture}
            \node[state, initial] (q00) {\tiny $(0,0)$};
            \node[state, above right of=q00] (q01) {\tiny $(0,1)$};
            \node[state, above right of=q01] (q02) {\tiny $(0,2)$};
            
            \node[state, below right of=q01] (q10) {\tiny $(1,0)$};
            \node[state, above right of=q10] (q11) {\tiny $(1,1)$};
            \node[state, above right of=q11] (q12) {\tiny $(1,2)$};
            
            \node[state, below right of=q11] (q20) {\tiny $(2,0)$};
            \node[state, above right of=q20] (q21) {\tiny $(2,1)$};
            \node[state, above right of=q21] (q22) {\tiny $(2,2)$};
    
            \node[state, accepting, below right of=q21] (q30) {\tiny $(3,0)$};
            
            
    
            \draw (q00) edge[above] node{$\epsilon$} (q10)
                  (q00) edge[above left] node{$\Sigma$} (q01)
                  (q01) edge[above right] node{$\epsilon$} (q10)
                  (q01) edge[above left] node{$\Sigma$} (q02)
                  (q02) edge[right] node{$\epsilon$} (q10)
                  
                  (q10) edge[above] node{$\epsilon$} (q20)
                  (q10) edge[above left] node{$\Sigma$} (q11)
                  (q11) edge[above right] node{$\epsilon$} (q20)
                  (q11) edge[above left] node{$\Sigma$} (q12)
                  (q12) edge[right] node{$\epsilon$} (q20)
                  
                  (q20) edge[above] node{$\epsilon$} (q30)
                  (q20) edge[above left] node{$\Sigma$} (q21)
                  (q21) edge[above right] node{$\epsilon$} (q30)
                  (q21) edge[above left] node{$\Sigma$} (q22)
                  (q22) edge[right] node{$\epsilon$} (q30)
                  
                ;
        \end{tikzpicture}
        }%
        \caption{$k$-constrained label and frame dependent alignment lattice with $T=3, k=2$.}
        \label{fig:k-constrained-label-and-frame-dependent}
    \end{subfigure}
    \hfill
    \begin{subfigure}[b]{0.45\textwidth}
        \centering
        \scalebox{0.75}{\begin{tikzpicture}
            \node[state, initial] (q0) {0};
            \node[state, right of=q0] (q1) {1};
            \node[state, right of=q1] (q2) {2};
            \node[state, accepting, right of=q2] (q3) {3};
    
            \draw (q0) edge[above] node{$\Sigma$} (q1)
                  (q0) edge[bend right, above] node{$\epsilon$} (q3)
                  (q1) edge[above] node{$\Sigma$} (q2)
                  (q1) edge[bend right, above] node{$\epsilon$} (q3)
                  (q2) edge[above] node{$\Sigma$} (q3)
                  (q2) edge[bend right, above] node{$\epsilon$} (q3);
        \end{tikzpicture}}
        \caption{Label dependent alignment lattice with $l(T) = 3$.}
        \label{fig:label-dependent}
    \end{subfigure}
    \caption{Examples of GNAT components}
\end{figure}
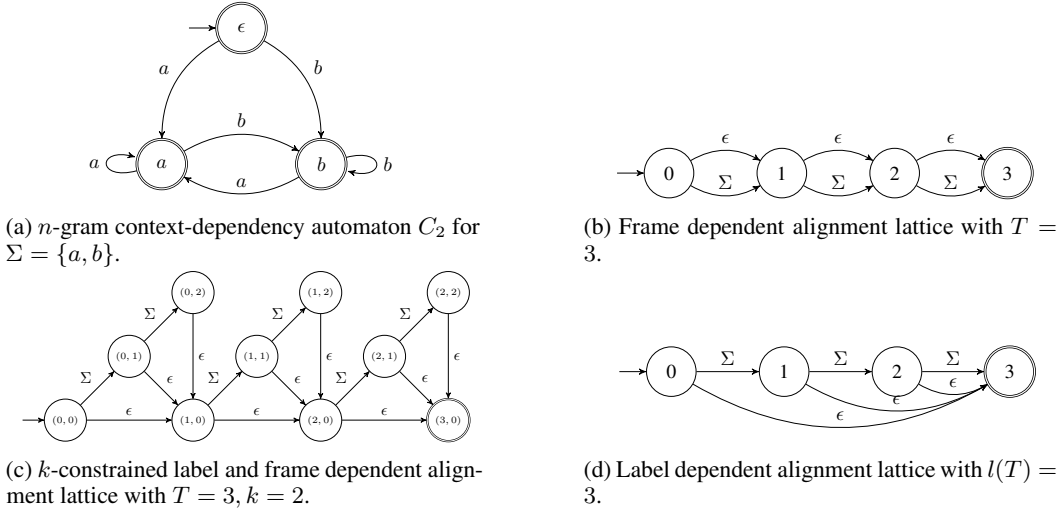

\subsection{Context Dependency} \label{subsec:context-dependency}

GNAT uses an $n$-gram
context-dependency defined by
$C_n = (\Sigma, Q_n, i_n, F_n, E_n)$, where $Q_n = \Sigma^{\leq n-1}$ corresponds to a
label history of length up to $n-1$. The initial state $i_n=\epsilon$ is the
empty label sequence. The transitions $E_n \subseteq Q_n \times \Sigma \times Q_n$ correspond to
truncated concatenation: $E_n = \bigl\{(q, y, q')\,|\, q \in Q, y \in \Sigma)\bigr\}$ where
$q'$ is the suffix of $qy$ with length at most $n - 1$. For
example, when $n=3$, the transition $(ab, c, bc)$ goes from state $ab$ to state $bc$ with label
$c$. All states are final: $F_n=Q_n$.
See Figure~\ref{fig:n-gram-context} for the FSA $C_2$ when $\Sigma = \{ a, b \}$.
The intersection $L_T \cap C_n$ is easy to compute thanks to the absence of $\epsilon$-transitions in $C_n$.
Appendix~\ref{sec:accelerator_friendly_modeling} demonstrates how $C_n$ can be efficiently intersected during the shortest distance computation.

Although not studied in this paper, our modular framework makes it easy to switch to a more sophisticated context dependency, such as clustered histories often used for context-dependent phone models,
or a variable context length as used in $n$-gram language models \cite{mohri2008}.

\subsection{Alignment Lattices} \label{subsec:alignment-lattice}

Given the feature sequence length $T$, the alignment lattice FSA $L_T$ defines all the possible alignments between the feature sequence and allowed label sequences.
Since the feature sequence length $T$ usually differs from the label sequence length $U$, many different alignments between the feature sequence and a label sequence can be defined.
The states in an alignment lattice FSA encode how the next label or $\epsilon$-transition corresponds to some position in the feature sequence.

We can choose different structures for $L_T$ by encoding one or both of the positions in the feature sequence and the label sequence.
A simple example is the \emph{frame dependent alignment} similar to \cite{graves2006connectionist}, where each frame is aligned to at most one label:
\begin{align*}
    Q_T & = \{ 0, \dots, T \} \\
    i_T & = 0 \\
    F_T & = \{ T \} \\
    E_T & = \bigl\{ (t-1, y, t) \mid y \in \Sigma \cup \{ \epsilon \},\ 1 \leq t \leq T \bigr\}
\end{align*}
Here any state $t < T$ represents a position in the feature sequence.
We start by aligning to the initial frame of the feature sequence, and repeatedly shift to the next frame for every subsequent label or $\epsilon$-transition until all frames have been visited.
Figure~\ref{fig:frame-dependent} depicts a frame dependent alignment lattice.

To allow a label sequence longer than the feature sequence, we can use the \emph{$k$-constrained label and frame dependent alignment} similar to \cite{graves2012sequence}:
\begin{align*}
    Q_T & = \bigl\{ (t, n) \mid 0 \leq t \leq T-1,\ 0 \leq n \leq k \bigr\} \cup \bigl\{ (T, 0) \bigr\} \\
    i_T & = (0, 0) \\
    F_T & = \bigl\{ (T, 0) \bigr\} \\
    \begin{split}
        E_T & = \Bigl\{ \bigl((t, n-1), y, (t, n)\bigr) \mid \\ & \qquad y \in \Sigma,\ 0 \leq t \leq T-1,\ 1 \leq n \leq k \Bigr\} \\ 
            & \qquad \cup \Bigl\{ \bigl((t-1, k), \epsilon, (t, 0)\bigr) \mid 1 \leq t \leq T \Bigr\}
    \end{split}
\end{align*}
Here, up to $k$ consecutive label transitions can align to any single frame.
An $\epsilon$-transition is then taken to explicitly shift the alignment to the next frame.
The number of labels aligned to one frame is constrained by a constant $k$ solely in order to impose acyclicity.
Figure~\ref{fig:k-constrained-label-and-frame-dependent} depicts a $k$-constrained label and frame dependent alignment lattice.

Some models may only depend on the position in the label sequence, similar to \cite{chan2015listen}.
In this case we can bound the length of the label sequence by some function $l(T)$, and use the following \emph{label dependent alignment}:
\begin{align*}
    Q_T & = \bigl\{ 0, \ldots, l(T) \bigr\} \\
    i_T & = 0 \\
    F_T & = \{ l(T) \} \\
    \begin{split}
        E_T & = \bigl\{ (u-1, y, u) \mid y \in \Sigma,\ 1 \leq u \leq l(T) \bigr\} \\
            & \qquad \cup \bigl\{ (u, \epsilon, l(T)) \mid 0 \leq u \leq l(T) - 1 \bigr\}
    \end{split}
\end{align*}
Here, each label can be seen as aligning to the entire feature sequence, and the $\epsilon$-transition serves as an explicit termination of the label sequence.
Figure~\ref{fig:label-dependent} depicts a label dependent alignment lattice.

\subsection{Weight functions} \label{subsec:weight-functions}

The weight function $\omega_{\theta,\vb{x}}$ translates trainable parameters into transition weights for states in the recognition lattice $A_{\theta,\vb{x}}$.
The choice of the weight function depends on the choice of the context and alignment lattice FSA, especially the alignment lattice where the meaning of a state directly affects how the weight function can access $\vb{x}$.

In the experiments discussed in this paper, we need concrete weight functions for frame dependent and $k$-constrained label and frame dependent alignment lattices.
In these two types of alignment lattices, a non-final state $q_a$ in $Q_T$ contains a position $\tau(q_a)$ in the feature sequence (the state itself in the case of frame dependent alignment lattices; the first value in the state in the case of label and frame dependent alignment lattices).
Weight functions can thus be defined in three steps,
\begin{compactenum}
    \item Feed $\vb{x}$ into an encoder, such as unidirectional or bidirectional RNN, or a self-attention encoder to obtain the sequence of hidden units $\vb{h}$ of dimension $D$. In the experiments we compared streaming vs non-streaming encoders.
    \item Map a single frame of hidden units $\vb{h}[t]$ and context state $q_c$ to a $(|\Sigma|+1)$-dimensional vector, corresponding to the unnormalized transition weights for $y \in \Sigma \cup \{ \epsilon \}$.
    \item Optionally locally normalize transition weights across $y \in \Sigma \cup \{ \epsilon \}$ given $(q_a, q_c)$.
\end{compactenum}

For step 2, we experiment with the following concrete modelling choices with varying degree of parameter sharing,

\textbf{Per-state linear projection} (\emph{unshared}) For every context state $q_c$, we obtain a $D \times (|\Sigma|+1)$ projection matrix $W_{q_c}$ and a $(|\Sigma|+1)$-dim bias vector $b_{q_c}$ from $\theta$, and define for $y \in \Sigma \cup \{ \epsilon \}$:
\[
\omega_{\theta,\vb{x}}(q_a, q_c, y) = \bigl(W_{q_c} \cdot \vb{h}[\tau(q_a)] + b_{q_c}\bigr)[y]
\]
\textbf{Shared linear projection with per-state embedding} (\emph{shared-emb}) We obtain from $\theta$ 
    \begin{inparaenum}[(a)]
        \item for every context state $q_c$ a $D$-dimensional state embedding $E_{q_c}$, 
        \item independent of context states a $D \times (|\Sigma|+1)$ projection matrix $W$ and a $(|\Sigma|+1)$-dim bias vector $b$
    \end{inparaenum}, and define for $y \in \Sigma \cup \{ \epsilon \}$:
\[
\omega_{\theta,\vb{x}}(q_a, q_c, y) = \bigl(W \cdot \tanh(\vb{h}[\tau(q_a)] + E_{q_c}) + b\bigr)[y]
\]
\textbf{Shared linear projection with RNN state embedding} (\emph{shared-rnn}) Similar to \emph{shared linear projection with per-state embedding} but $E_{q_c}$ is obtained from running an RNN (e.g. LSTM) on the $n$-gram label sequence represented by $q_c$.

\section{Discussion}
\label{sec:discussions}

\textbf{A Modular Framework \quad}
All the existing locally normalized models can be explained within the modular framework
presented in Section~\ref{sec:gnat-model}
with a particular choice of context size, alignment lattice
and of course with constraining the weights to be locally normalized using softmax function.
Appendix~\ref{sec:universality} presents how CE, CTC, LAS, RNNT and HAT models
can be expressed within this framework. This allows controlled comparison of
different components as well as creating new models by mixing different
modeling choices.
Like locally and globally models in general, when the weights are locally normalized, the denominator of models defined in our framework is one.

Finally, our framework is very different from traditional uses of finite state machines via a cascade of weight finite state transducer compositions.
Our separation of the weight function from the automaton topology allows an arbitrarily complex, non-linear weight function to model the dependency among alignment states, context states, and output label, which is impossible with composition cascades.

\textbf{Related Globally Normalized Models\quad}
There is a rich literature on the applications of globally normalized
models \cite{bottou1997global, lecun1998gradient, rosenfeld1997whole, lafferty2001conditional} as
well as detailed studies on the importance of global normalization in addressing the label bias problem
\cite{lafferty2001conditional, andor2016globally, goyal2019empirical}.
In the context
of ASR there is a lot of research on applying globally normalized models \cite{bahl1986maximum, brown1987acoustic, bridle1991alphanet, gunawardana2005hidden, layton2007augmented, zweig2009segmental, hifny2009speech}. Among these, MMI \cite{bahl1986maximum, brown1987acoustic} is the most relevant
globally normalized criterion to
our work. In MMI a sequence level score is factorized by a likelihood score
which comes from an acoustic model and a prior score which is usually a word based
language model (LM). The denominator score is then approximated over a lattice
of hypotheses. More recently a lattice free version of this criterion has been
introduced \cite{povey2016purely} which replaces the word level LM with a 4-gram
phone LM.
Both the GNAT and MMI criteria are globally normalized.
The GNAT model differentiates itself from MMI in several ways. First unlike MMI it
does not require an external LM and it does not apply any constraint on how the sequence level
scores are defined. Second, in MMI the language model is kept frozen while the acoustic model parameters
are updated via optimization of the MMI criterion. In GNAT all the model parameters
are trained together.
Third, GNAT provides the exact computation
of the denominator while standard MMI only offers an approximation. Finally GNAT trains from scratch
without any need for initialization or special regularization techniques as used in the
lattice-free version of MMI \cite{povey2016purely}. In addition we were able to train
GNAT models with accelerators without any techniques discussed in \cite{povey2016purely}.

Finally the concept of global normalization has also been visited with deep neural networks
\cite{hannun2020differentiable, collobert2016wav2letter, xiang2019crf, zheng2021advancing}. 
These models can be seen as special cases of MMI thus all the differences between MMI and
GNAT model applies here as well.
Apart from their weaker modelling power as a result of using WFST composition cascades, all these models are non-streaming, where as
explained multiple times in our paper, global normalization and local normalization
are equally expressive.

\textbf{Challenges \quad}
The main challenge with the GNAT model is its scalability to a larger number of label contexts.
At each training step, the model requires $2 \times |Q_C| \times |\Sigma|$ multiplication
and summation. For $n$-gram context dependency, $|Q_c| = \frac{|\Sigma|^{n+1} - 1}{|\Sigma| - 1}$,
thus the computation scale exponentially by value of $n$. However as shown in Appendix~\ref{sec:appendix_c},
due to the particular structure of this space the practical computation and memory cost benchmarks do not scale exponentially
with $n$. We also note that large value of $n$
might also not be necessary:
The HAT model \cite{variani2020hybrid}
reports that a Seq2Seq model with a label history of just the two previous
phonemes performs on par with a similar model with a full history trained on very large voice-search corpus. Similar observations
are reported in studies with grapheme and wordpiece units \cite{zhang2020transformer, ghodsi2020rnn}.
Due to the data sparsity there might not be enough training to fully represent
a $n$-gram space, so increasing the value of $n$ might not necessarily lead to performance
improvement.
One way of dealing with large number of states is to use standard pruning techniques to keep
only some of the most common states in the training data.

\section{Experiments}
\label{sec:experiments}

\textbf{Data \quad}
We use the full $960$-hour Librispeech corpus \cite{panayotov2015librispeech} for experiments.
The input features are 80-dim. log Mel extracted
from a 25 ms window of the speech signal with a 10 ms shift.
The SpecAugment library with baseline recipe parameters were used \cite{gulati2020conformer}.
The transcript
truth is used without any processing and tokenized by the $28$ graphemes
that appear in the training data.

\textbf{Architecture \quad}
Attention-based architectures allows using the same parameterization 
for streaming and non-streaming models, thus for 
all the experiments we used $12$-layer Conformer encoders \cite{gulati2020conformer} with
model dimension 512, followed by a linear layer with output dimension 640.
The Conformer parameters are set such that the only difference between streaming
and non-streaming models is the right context:
at each time frame $t$, the streaming models only access the left context (feature frames from $1$ to $t$), while the non-streaming models can see the entire
acoustic feature sequence. To enforce the consistency of the encoder architecture
between streaming and non-streaming modes, we removed all the sub-architecture which
behaved differently between these two modes. Specifically, we removed the convolution
sub-sampling layer, and also forced the stacking layers to only stack
within the left context.
The baseline experiments use a shared-rnn weight function defined in
section~\ref{subsec:weight-functions}. A single layer LSTM is used with 640 cells.
The experiments with the unshared weight function use a linear layer of size
$(|Q_C| \times  |\Sigma|) \times 640$ to project the encoder activation at each time
frame into the transition weights of the recognition lattice. In our experiments,
$|\Sigma|=32$. For the $n$-gram context 
dependency, $|Q_c| = \frac{|\Sigma|^{n+1} - 1}{|\Sigma| - 1}$. The experiments
with the shared-emb weight function use an embedding table of size $|Q_c| \times 128$. 

\textbf{Training \quad}
All models are trained on $8\times8$ TPUs with a batch size $2048$.
The training examples with more than $1961$ feature frames or more than
$384$ labels are filtered out. We used Adam optimizer
\cite{kingma2014adam} ($\beta_1=0.9$, $\beta_2=0.98$,
and $\epsilon=10^-9$) with the Transformer learning rate schedule 
\cite{vaswani2017attention} (10k warm-up steps and peak learning 
rate $0.05/\sqrt{512}$). We applied the same regularization techniques
and the training hyperparameters used in the baseline recipe of \cite{gulati2020conformer}.

\textbf{Evaluation \quad}
We report WER results on standard Librispeech test
sets: test\_clean and test\_other.
The WER is either computed with \emph{sum-path} algorithm or \emph{max-path} algorithm. The
sum-path algorithm merges the alignment hypothesis corresponding
to the same label sequence prefix after removal of epsilons. 
In ideal decoding, sum-path should result in the most likely output label sequence.
The max-path algorithm computes the highest scoring path using
algorithms in Appendix~\ref{sec:accelerator_friendly_modeling}.

\textbf{Baselines \quad}
The RNN-T baselines are presented in the row corresponding to the $M$-gram context dependency in Table~\ref{tab:local-baselines}.
For frame dependent alignment lattice, $M$ is equal to the length of the longest feature sequence in the training data ($1961$) and for label frame dependent $M$
is equal to $1961 + 384$, sum of the maximum feature sequence length and the maximum label
sequence length. The label frame dependent alignment lattice used for the baseline and all the other experiments is $k$-constrained with $k$ set to the number of
labels for each training example (maximum value 384).
The WER difference between
non-streaming baselines in Table~\ref{tab:local-baselines} and \cite{gulati2020conformer}
are mainly due to our modifications to the Conformer encoder for a controlled comparison against streaming models.

\begin{table}
    \addtolength{\leftskip} {-1em}
    \addtolength{\rightskip}{-1em}
    \begin{subtable}[b]{0.5\textwidth}\scalebox{0.9}{%
        \centering
        \begin{tabular}[t]{|c|c|c|c|c|}
        \hline
        context    & alignment  & weight fn & \multicolumn{2}{c|}{WER [\%]} \\
        \cline{4-5}
        dep. &  lattice   &    streaming       & clean & other \\
        \hline
        \multirow{4}{*}{$0$-gram} & \multirow{2}{*}{frame} & no & 4.0 & 10.0\\
        \cline{3-5}
                            &                              & yes & 7.1 & 16.0 \\
        \cline{2-5}
                            & \multirow{2}{*}{label frame} & no & 6.7 & 10.2\\
        \cline{3-5}
                            &                              & yes & 8.8 & 14.5\\
        \hline
        \multirow{4}{*}{$1$-gram} & \multirow{2}{*}{frame} & no & 2.8 & 6.0\\
        \cline{3-5}
                            &                              & yes & 4.9 & 10.0\\
        \cline{2-5}
                            & \multirow{2}{*}{label frame} & no & 2.5 & 5.6\\
        \cline{3-5}
                            &                              & yes & 5.1 & 10.3\\
        \hline
        \multirow{4}{*}{$2$-gram} & \multirow{2}{*}{frame} & no & 2.5 & 5.3\\
        \cline{3-5}
                            &                              & yes & 4.9 & 9.7\\
        \cline{2-5}
                            & \multirow{2}{*}{label frame} & no & 2.5 & 5.3\\
        \cline{3-5}
                            &                              & yes &5.0 & 9.8 \\
        \hline
        \multirow{4}{*}{M-gram} & \multirow{2}{*}{frame}   & no & 2.5 & 5.3\\
        \cline{3-5}
                            &                              & yes & 5.1 & 9.8\\
        \cline{2-5}
                            & \multirow{2}{*}{label frame} & no & 2.5 & 5.5\\
        \cline{3-5}
                            &                              & yes &5.0 & 9.8\\
        \hline
        \end{tabular}
        }%
        \caption{\label{tab:local-baselines} Locally normalized baselines with different context-dependency and alignment lattice. All models used a shared-rnn weight function with or without a streaming encoder. (sum-path decoding)}
    \end{subtable}
    \hfill
    \begin{subtable}[b]{0.5\textwidth}
        \centering
        \begin{subtable}[b]{\textwidth}\scalebox{0.9}{%
            \centering
            \begin{tabular}[t]{|c|c|c|c|c|c|}
            \hline
            context    & \multicolumn{2}{c|}{weight function}    & \multicolumn{2}{c|}{WER [\%]} \\
            \cline{2-5}
            dep. &    streaming       &   normalization             & clean & other \\
            \hline
            \multirow{4}{*}{1-gram} & \multirow{2}{*}{no}  & local  & 3.4 & 8.7 \\
            
            \cline{3-5}
                               &                      & global & 3.3 & 8.4 \\
            \cline{2-5}
                               & \multirow{2}{*}{yes} & local  & 7.0 & 17.4 \\
            \cline{3-5}
                               &                      & global & 5.5 & 14.0 \\
            \hline
            \multirow{4}{*}{2-gram} & 
            \multirow{2}{*}{no}  & local  & 2.8 & 6.7 \\
            \cline{3-5}
                               &                      & global & 2.8 &  6.7 \\
            \cline{2-5}
                               & 
            \multirow{2}{*}{yes} & local  & 4.9 & 11.0 \\
            \cline{3-5}
                               &                      & global & 3.8 & 9.5 \\
            \cline{2-5}
            \hline
            \end{tabular}
            }%
            \caption{\label{tab:global-local} Weight function parameters: normalization and streaming. (max-path decoding)}
        \end{subtable}
        \begin{subtable}[b]{\textwidth}\scalebox{0.9}{%
            \centering
            \begin{tabular}{|c|c|c|c|}
            \hline
             \multicolumn{2}{|c|}{weight function} & \multicolumn{2}{c|}{WER [\%]} \\
             \hline
             type & normalization & clean & other \\
             \hline
             \multirow{2}{*}{unshared} & local & 4.9 & 10.7\\
            \cline{2-4}
                                       & global & 4.2 & 10.6 \\
            \hline
             \multirow{2}{*}{shared-emb} & local & 5.4 & 13.1\\
            \cline{2-4}
                                        &  global & 4.1 & 9.9 \\
            \hline
             \multirow{2}{*}{shared-rnn} & local & 4.9 & 11.0\\
            \cline{2-4}
                                       &    global & 3.8 & 9.5\\ 
            \hline
            \end{tabular}
            }%
            \caption{\label{tab:weight_functions} Comparison of different weight function types (max-path decoding).}
        \end{subtable}
    \end{subtable}
    \caption{Experiment results}
    \vspace{-1.0cm}
\end{table}

\textbf{Choice of the context dependency \quad}
Table~\ref{tab:local-baselines} compares
effect of $n$-gram context dependency for $n=0,1,2$ and baseline RNN-T models. The general observation is that increasing $n$ leads to better performance quality
independent of the other choices of the modeling parameters. However, the
model with $2$-gram context dependency already performs on par with RNN-T baseline. The $1$-gram context dependency perform almost on par as baseline on clean test set while still lagging on the other test set. This is consistent with the earlier observations
in \cite{variani2020hybrid, zhang2020transformer, ghodsi2020rnn}.

\textbf{Choice of the alignment lattice \quad}
The comparison of different alignment lattices in Table~\ref{tab:local-baselines}
suggests that this choice does not significantly contribute to the model performance.
While there is a performance gap for $0$-gram context dependency, we do not think there is a principal argument in favor of frame dependent alignment lattice. We speculate that this is more due to the choice of optimization parameters.
However, the choice of lattice type can have some side effects. For example as $k$ in $k$-constrained label frame dependent alignment lattice increases, the model has more ability to delay its prediction to the end of the signal.
This implicit lookahead can translate into performance gains particularly for unidirectional models. By limiting this quantity to 1, we observed that performance on clean and other sets degrades by $34.9 \%$ and $36.7 \%$, respectively.

\textbf{Choice of the weight function normalization \quad}
Table~\ref{tab:global-local} examines the effect of weight normalization on non-streaming and streaming models. Here we present models with $1$-gram and
$2$-gram context dependency with a frame dependent alignment lattice. Note
that for $0$-gram context dependency with a frame dependent alignment lattice it is
easy to show that locally normalized and globally normalized models are equivalent.
For non-streaming models, the normalization seems to not have an impact on the performance
quality neither for $1$-gram nor for $2$-gram context dependency experiments.
This is expected since the full acoustic feature sequence context allows the model
to avoid the label bias problem \cite{andor2016globally, goyal2019empirical} which is consistent with the equal expressiveness
of globally and locally normalized models under non-streaming setting \cite{smith2007weighted}.
On the other hand, streaming models significantly benefit from global normalization:
For clean test set, the globally normalized model outperforms the locally normalized
model by about $21 \%$ relative WER for $1$-gram context dependency and by about $20 \%$
relative gain for $2$-gram context dependency.



The globally normalized model with $2$-gram context dependency also beat the baseline
streaming RNN-T model in Table~\ref{tab:local-baselines} and performs significantly closer
to the non-streaming RNN-T baseline. The equivalent streaming RNN-T model performs $5.1 \%$
on test clean and the non-streaming model performs $2.5 \%$ on same test set. The globally
normalized model decoded with max-path algorithm performs $3.8 \%$ on same test set. The globally normalized
model effectively closed almost $50 \%$ of the performance gap between streaming and non-streaming models.

The reported performance for the globally normalized models is from max-path decoding, while the baselines benefit from sum-path decoding. 
Comparing the locally normalized models' WER from max-path decoding
in Table~\ref{tab:global-local} and their counterparts in Table~\ref{tab:local-baselines},
it is clear that sum-path decoding leads to an extra WER gain.
This gain is more significant on test\_other.
So we expect the globally model performs even better when decoded with sum-path. 

The standard sum-path algorithms use several heuristics particularly for path merging and pruning.
While similar merging techniques can be applied to the globally normalized models, the pruning heuristics 
require several adjustments. This is particularly due to the nature of the globally normalized models where
the transition weights are not constrained and can take any value, unlike locally normalized models where
the transition weights are constrained to be positive number between $0$ and $1$ and sum to $1$ for all the weights
leaving the same state in recognition lattice. We can also reduce the performance gap between max-path and sum-path
by constraining the training criterion to distribute the whole probability mas into one one alignment path. This can be done by constraining the objective function with alignment path entropy. This effectively avoids the need for sum-path inference. We
will present these approaches in our future publication.

\textbf{Choice of the weight function architecture\quad}
Finally Table~\ref{tab:weight_functions} compares different choices of the architectures for a streaming model
with $2$-gram context dependency and frame dependent alignment lattice. 
While the unshared and shared-rnn architectures are very different in terms of parameter sharing among
states, both perform well, though the shared-rnn architecture performs slightly better. The shared-emb
architecture performs significantly worst than shared-rnn architecture. Note that
the shared-rnn model is able to learn common structures across states 
in the context dependency while shared-emb does not have such capability.

\vspace{-0.25cm}
\section{Conclusion}
\label{sec:conclusion}

The GNAT model was proposed and evaluated with the focus on the label
bias problem and its impact on the performance gap between streaming and non-streaming
locally normalized ASR.
The finite context property of this model allows exact
computation of the sequence level normalization which makes this model differ from existing
globally normalized models. Furthermore, the same property allows
accelerator friendly training and inference.
We showed that the streaming models with globally normalized
criteria can significantly close the gap between streaming and non-streaming
models by more than 50\%.
Finally, the modular framework introduced in this paper to explain the GNAT model encompasses all
the common neural speech recognition models. This enables fair and accurate comparison of different
models via controlled modelling choices and creation of new ASR models.

\newpage
\bibliographystyle{plain}
\bibliography{gnat}

\newpage
\appendix
\onecolumn
\section{An overview example}
\label{sec:appendix_a}

To illustrate different components of the GNAT model, here we present a toy example 
of designing a speech recognition for finite alphabet $\Sigma = \{a, b\}$.
Given an input feature sequence $\vb{x} = x_1 \dots x_T$, we wish to predict the
corresponding output label sequence $\vb{y} = y_1 \dots y_U,\ y_i \in \Sigma$.
Our objective is to create a conditional probabilistic model $P(\vb{y} | \vb{x})$
which assigns the highest probability to the correct label sequences for any given feature
sequence. We construct a GNAT model with the following components:
\begin{itemize}
    \item context-dependency: $2$-gram
    \item alignment-lattice: frame dependent
    \item weight function: per-state linear projection, streaming
\end{itemize}
Next we elaborate details of each of these modules and how they are
integrated to create the final space the recognition lattice $A_{\theta,\vb{x}}$ and the probabilistic
model $P(\vb{y} | \vb{x})$ as described in Section~\ref{sec:gnat-model}.

\subsection{Context Dependency FSA}
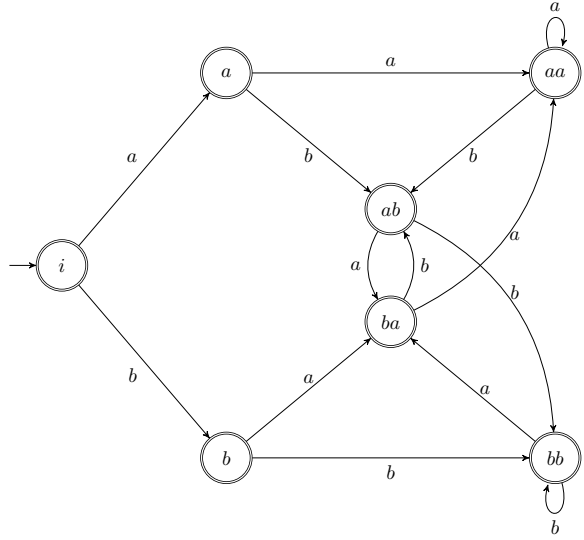
\begin{figure}[h]
    \centering
    \scalebox{0.75}{%
    \begin{tikzpicture}
        \node[state, initial, accepting] (empty) {$i$};
        \node[state, accepting, above right of=empty, xshift=1.5cm, yshift=2.0cm] (a) {$a$};
        \node[state, accepting, below right of=empty, xshift=1.5cm, yshift=-2.0cm] (b) {$b$};
        \node[state, accepting, below right of=a, xshift=1.5cm, yshift=-1.0cm] (ab) {$ab$};
        \node[state, accepting, above right of=ab, xshift=1.5cm, yshift=1.0cm] (aa) {$aa$};
        \node[state, accepting, above right of=b, xshift=1.5cm, yshift=1.0cm] (ba) {$ba$};
        \node[state, accepting, below right of=ba, xshift=1.5cm, yshift=-1.0cm] (bb) {$bb$};

        \draw  (empty) edge[above left] node{$a$} (a)
               (empty) edge[below left] node{$b$} (b)
               (a) edge[right, above] node{$a$} (aa)
               (a) edge[below] node{$b$} (ab)
               (b) edge[right, below] node{$b$} (bb)
               (b) edge[above] node{$a$} (ba)
               (aa) edge[loop above, above] node{$a$} (aa)
               (aa) edge[below] node{$b$} (ab)
               (ab) edge[bend right, left] node{$a$} (ba)
               (ab) edge[bend left, above] node{$b$} (bb)
               (ba) edge[bend right, below] node{$a$} (aa)
               (ba) edge[bend right, right] node{$b$} (ab)
               (bb) edge[right] node{$a$} (ba)
               (bb) edge[loop below, below] node{$b$} (bb);
    \end{tikzpicture}
    }%
    \caption{2-gram context-dependency automaton $C_3$ for $\Sigma=\{a, b\}$.}
    \label{fig:2-gram-fst}
\end{figure}

Figure~\ref{fig:2-gram-fst} presents the $2$-gram context dependency $C_3$.
The set of states for this space are the initial state, 1-gram states and
2-gram states:
\[
Q_C = \bigl \{i, a, b, aa, ab, ba, bb\bigr \}
\]
With the lexicographic order, these states are
indexed as follow:

\begin{tabular}{ccc}
  state &	state index \\
  \hline
  i     &       0  \\
  a     &       1  \\
  b     &       2  \\
  aa    &       3  \\
  ab    &       4  \\
  ba    &       5  \\
  bb    &       6  \\
\end{tabular}

For this particular FSA, the transitions space is
\[
E_C = \bigl\{(q, y, q')\,|\, q \in Q, y \in \Sigma)\bigr\}
\]
where $q'$ is the suffix of $qy$ with length at most $2$.
All 14 transitions of this space are listed in the following table:

\begin{tabular}{ccc}
from state & label & to state \\
\hline
$i$ & $a$ & $a$ \\
$i$ & $b$ & $b$ \\
$a$ & $a$ & $aa$ \\
$a$ & $b$ & $ab$ \\
$b$ & $a$ & $ba$ \\
$b$ & $b$ & $bb$ \\
$aa$ & $a$ & $aa$ \\
$aa$ & $b$ & $ab$ \\
$ab$ & $a$ & $ba$ \\
$ab$ & $b$ & $bb$ \\
$ba$ & $a$ & $aa$ \\
$ba$ & $b$ & $bb$ \\
$bb$ & $a$ & $ba$ \\
$bb$ & $b$ & $bb$ \\
\end{tabular}

\subsection{Alignment Lattice FSA}

Figure~\ref{fig:frame-dependent} depicts a frame dependent alignment lattice $L_4$
for four frames feature sequence $\vb{x}$. The states of this space are:
\[
Q_T = \bigl \{0, 1, 2, 3, 4 \bigr \}
\]
where $0$ is the initial state and $4$ is the final state.
Every path starting from the initial state in this automaton corresponds to one
possible alignment sequence of the input feature sequence. The example FSA in
Figure~\ref{fig:frame-dependent} encodes $3 ^ 4 = 81$ possible alignment
sequences. An alignment path $\epsilon a \epsilon b$ corresponds to the following sequence of transitions in $L_4$:
\[
(0, \epsilon, 1), (1, a, 2), (2, \epsilon, 3), (3, b, 4)
\]

\subsection{Weight Function}

The weight function
$\omega_{\theta,\vb{x}}: Q_T \times Q_C \times (\Sigma \cup \epsilon) \rightarrow \mathbb{K}$. $\omega_{\theta,\vb{x}}$
is the only trainable component of the GNAT model which assigns a weight to every transition of the recognition lattice $A_{\theta,\vb{x}}$. We first feed $\vb{x}=x_1, ..., x_4$ into an
encoder to obtain hidden activations $\vb{h}=h_1, ..., h_4$ of dimension $D$. The encoder can be any neural architecture such as DNNs, CNNs, RNNs or Transformers. Since we are interested in streaming weight function for this example, we need to make sure the encoder is also streaming. This means $h_t$ can only depends on
$x_{1:t-1}$. Finally we define a $D \times 3$ matrix $W_{q_c}$ and a $3$-dim bias
vector $b_{q_c}$ for any $q_c \in Q_c$. The weight function is then defined as:
\[
\omega_{\theta, x_{1:t}}(q_a=t, q_c, y) = W_{q_c}[y, :] \cdot h_t + b_{q_c} [y]
\]
where $W_{q_c}[y, :]$ is the row of $W_{q_c}$ corresponding to label $y \in \Sigma \cup \{ \epsilon \}$.
The total number of trainable parameters is $7 \times D \times 3 + 7 \times 3$ parameters plus the number
of parameters of the encoder function.

\subsection{The recognition lattice $A_{\theta,\vb{x}}$}

Given the context dependency FSA $C_3$, the frame dependent alignment lattice $L_4$ and the weight function $\omega_{\theta,\vb{x}}$, we are ready to derive the recognition lattice $A_{\theta,\vb{x}}$. The state space has $5 \times 7$ states:
\[
Q_{\theta, \vb{x}} = \bigl \{ (t, q_c) \, | \, 0 \leq t \leq 4, q_c \in Q_c \bigr \}
\]
The transitions in this space is specified by the state it is originated from, $(t, q_c)$, the label $y \in \Sigma \cup \{ \epsilon\}$, weight $\omega_{\theta, \vb{x}}(t, q_c, y)$ and the state the transition is ended to $(t+1, q_c')$. For the alignment
sequence of our example, $\epsilon a \epsilon b$, the transitions are:

\begin{tabular}{cccc}
  from    & label      &  weight & to \\
  \hline
  $(0, i)$      & $\epsilon$ & $\omega_{\theta, x_1}(q_a=0, q_c=i, y=\epsilon) = W_0[0, :] \cdot h_1 + b_0[0]$ & $(1, \epsilon)$ \\
  $(1, i)$      & $a$        & $\omega_{\theta, x_{1:2}}(q_a=1, q_c=i, y=a) = W_0[1, :] \cdot h_2 + b_0[1]$  & $(2, a)$\\
  $(2, a)$      & $\epsilon$ & $\omega_{\theta, x_{1:3}}(q_a=2, q_c=a, y=\epsilon) = W_1[0, :] \cdot h_3 + b_1[0]$ & $(3, a)$ \\
  $(3, a)$      & $b$        & $\omega_{\theta, x_{1:4}}(q_a=3, q_c=a, y=b) = W_1[2, :] \cdot h_4 + b_1[2]$ & $(4, ab)$\\
  \hline
\end{tabular}

The product of the above weights is the score that the GNAT model assigns to the features sequence $\vb{x}$ and alignment sequence $\epsilon a \epsilon b$:
\[
\text{score}(\vb{x},\epsilon a \epsilon b ) = \omega_{\theta, x_1}(0, i, \epsilon) \omega_{\theta, x_{1:2}}(1, i, a) \omega_{\theta, x_{1:3}}(2, a, \epsilon) \omega_{\theta, x_{1:4}}(3, a, b)
\]
Note that for simplicity we use real semiring for all the score calculations in this example.

The GNAT model formulates the posterior probability $P(\vb{y} | \vb{x})$
by ratio of two quantities:
\begin{itemize}
    \item numerator: sum of all the $\text{score}(\vb{x}, \tilde{\vb{y}})$ where
    $\tilde{\vb{y}}$ is an alignment between $\vb{x}$ and $\vb{y}$. For example
    if $\vb{y} = a b$, there are only 6 possible alignments: $a b \epsilon \epsilon$, $a \epsilon b \epsilon$, $a \epsilon \epsilon b$, $\epsilon a b \epsilon$, $\epsilon a \epsilon b$, $\epsilon \epsilon a b$
    \item denominator: sum of all the $\text{score}(\vb{x}, \tilde{\vb{y}})$ where
    $\tilde{\vb{y}}$ can be any $(\Sigma + 1)^T = 3^4=81$ sequences.
\end{itemize}
Since the numerator computation is a special case of the denominator, we only present
the denominator calculation. To follow the computation presented in
Section~\ref{sec:accelerator_friendly_modeling}, we first define forward variable
$\alpha_t$ which is a $7$-dim real-valued vector where $\alpha_t [j]$ is the total alignment scores
reaching to state index $j$ (corresponding to the state indices in $Q_c$) at time $t$:
\[
\alpha_t =
\begin{blockarray}{ccccccc}
 (t, i) & (t, a) & (t, b) & (t, aa), & (t, ab) & (t, ba) & (t, bb)\\
\begin{block}{(ccccccc)}
 \alpha_t[0] & \alpha_t[1] & \alpha_t[2]  & \alpha_t[3] & \alpha_t[4] & \alpha_t[5] & \alpha_t[6]\\
\end{block}
\end{blockarray}
\]
the initial state $\alpha_0$ is a $1$-hot vector with $\alpha_0[j]=1$ iff $j=0$,
the initial state. At every time frame $t$, the transition weight matrix 
$\Omega_{t, \Sigma}$ is defined for all the transitions in $E_{A_{\theta,\vb{x}}}$
where label is an element of $\Sigma$.
This matrix is a structured matrix with only $|\Sigma|=2$ non-zero elements per row:
\[
\Omega_{t, \Sigma} = 
\begin{blockarray}{cccccccc}
& (t, i) & (t, a) & (t, b) & (t, aa), & (t, ab) & (t, ba) & (t, bb)\\
\begin{block}{c(ccccccc)}
  (t-1, i)  & 0 & \omega(t, i, a) & \omega(t, i, b)  & 0 & 0 & 0 & 0\\
  (t-1, a)  & 0 & 0 & 0 & \omega(t, a, a) & \omega(t, a, b) & 0 & 0\\
  (t-1, b)  & 0 & 0 & 0 & 0 & 0 & \omega(t, b, a) & \omega(t, b, b)\\
  (t-1, aa) & 0 & 0 & 0 & \omega(t, aa, a) & \omega(t, aa, b) & 0 & 0\\
  (t-1, ab) & 0 & 0 & 0 & 0 & 0 & \omega(t, ab, a) & \omega(t, ab, b)\\
  (t-1, ba) & 0 & 0 & 0 & \omega(t, ba, a) & \omega(t, ba, b) & 0 & 0\\
  (t-1, bb) & 0 & 0 & 0 & 0 & 0 & \omega(t, bb, a) & \omega(t, bb, b)\\
\end{block}
\end{blockarray}
\]
Similarly we denote $\Omega_{t, \epsilon }$ to be the transition weight matrix
for all the transitions in $E_{A_{\theta,\vb{x}}}$
where label is $\epsilon$. This matrix is a diagonal matrix corresponding to the weights
of the self loops:
\[
\Omega_{t, \epsilon } = 
\begin{blockarray}{cccccccc}
& (t, i) & (t, a) & (t, b) & (t, aa), & (t, ab) & (t, ba) & (t, bb)\\
\begin{block}{c(ccccccc)}
  (t-1, i)  & \omega(t, i, \epsilon) & 0 & 0 & 0 & 0 & 0 & 0\\
  (t-1, a)  & 0 & \omega(t, a, \epsilon) & 0 & 0 & 0 & 0 & 0\\
  (t-1, b)  & 0 & 0 & \omega(t, b, \epsilon) & 0 & 0 & 0 & 0\\
  (t-1, aa) & 0 & 0 & 0 & \omega(t, aa, \epsilon) & 0 & 0 & 0\\
  (t-1, ab) & 0 & 0 & 0 & 0 & \omega(t, ab, \epsilon) & 0 & 0\\
  (t-1, ba) & 0 & 0 & 0 & 0 & 0 & \omega(t, ba, \epsilon) & 0\\
  (t-1, bb) & 0 & 0 & 0 & 0 & 0 & 0 & \omega(t, bb, \epsilon)\\
\end{block}
\end{blockarray}
\]

For our model, the forward variable $\alpha_{t}$ can be calculated given
$\alpha_{t-1}$ and the above weights matrices as:
\[
\alpha_{t} = \alpha_{t-1}' (\Omega_{t, \Sigma} + \Omega_{t, \epsilon })
\]
since every transition at time $t+1$ is either an $\epsilon$ transition
or a non-$\epsilon$ transition.  Here $\alpha_{t-1}'$ is the transpose of
forward variable $\alpha_{t-1}$.

Given the above iterative equation, the
forward variable at time $4$ is equal to:
\[
\alpha_{4} = \alpha_0 (\Omega_{1, \Sigma} + \Omega_{1, \epsilon })(\Omega_{2, \Sigma} + \Omega_{2, \epsilon })(\Omega_{3, \Sigma} + \Omega_{3, \epsilon })(\Omega_{4, \Sigma} + \Omega_{4, \epsilon })
\]
and the denominator of the GNAT model is equal to $\sum_{j=0} ^ 6 \alpha_{4}[j]$.
Replacing the real semiring with tropical semiring in above calculation  will
allow us to find the most likely alignment sequence.
\section{A Modular Framework}
\label{sec:universality}

In this section we demonstrates how the existing and common neural speech recognition
models can be expressed within our proposed framework.

\subsection{Cross-entropy with Alignments}

The conventional cross-entropy models with feed-forward neural architectures \cite{morgan1990continuous} define the conditional probability of label
sequence $\vb{y}$ given feature sequence $\vb{x}$ by:
\[
P_\theta(\vb{y} | \vb{x}) = \prod_{t=1}^T P_\theta (y_t | x_t)
\]
where probability factors $P_\theta (y_t | x_t)$ are derived by some
neural architecture parameterized by $\theta$:
\[
P_\theta (y = y_t | x_t) = \frac{\exp( W[y_t, :] \cdot h_t + b[y_t])}{\sum_{y \in \Sigma} \exp( W[y, :] \cdot h_t + b[y])}
\]
where $h_t$ is the encoder activation of dimension $D$ at time frame $t$, $W$ is a
weight matrix of shape $|\Sigma| \times D$ and $b$ is a $|Sigma|$-dim bias vector.

The equivalent GNAT model is configured as follow:
\begin{itemize}
    \item context dependency: $0$-gram $C_1$
    \item alignment lattice: frame dependent without $\epsilon$ transitions
    \item weight function:
    \begin{itemize}
        \item $\omega_{\theta, \vb{x}}(q_a=t, q_c=i, y=y_t) \triangleq P_\theta (y = y_t | x_t)$
        \item locally normalized
        \item streaming
    \end{itemize}
\end{itemize}
here $i$ is initial state of $C_1$.

The more advanced cross-entropy models use recurrent architectures or transformers
as encoder \cite{variani2017end}. The only difference between the GNAT equivalent of these models and above configuration is that whether the encoder is
streaming or not.

\subsection{Listen, Attend and Spell (LAS)}

This model formulates the posterior probability by directly applying chain rule (Eq1 in \cite{chan2015listen}):
\[
P_{\theta} (\vb{y} | \vb{x}) = \prod_{l} P_{\theta} (y_l | \vb{x}, y_{< l})
\]
the posterior factors are defined as (Eq6-Eq8 of \cite{chan2015listen}):
\[
P_{\theta} (y_l | \vb{x}, y_{< l}) = \text{CharacterDistribution}(s_l
, c_l)
\]
where
\begin{eqnarray}
    \vb{h} &=& \text{Listen}(\vb{x}) \nonumber \\
    s_l &=& \text{RNN}(s_{l-1}, y_{l-1}, c_{l-1}) \nonumber \\
    c_l &=& \text{AttentionContext}(s_l, \vb{h}) \nonumber
\end{eqnarray}
here $\text{Listen}$ is a bidirectional encoder function, $\text{AttentionContext}$
is the attention network (Eq9-Eq11 of \cite{chan2015listen}).

The equivalent GNAT model is configured as follow:
\begin{itemize}
    \item context dependency: $M$-gram where $M$ is the length of longest label sequence in the training data. Note that $M$-gram context dependency is equivalent
    of the tree space truncated at depth $M$.
    \item alignment lattice: label dependent since the probability factorizes only on label sequence.
    \item weight function:
    \begin{itemize}
        \item $\omega_{\theta, \vb{x}}(q_a=l, q_c=q, y=y_l) \triangleq P_\theta (y = y_l | \vb{x}, y_{< l})$
        \item locally normalized
        \item non-streaming
    \end{itemize}
\end{itemize}

\subsection{Recurrent Neural Transducer}

The RNNT model formulate the posterior probability as marginalization of alignment sequences (Eq1 in the RNNT paper \cite{graves2012sequence}):
\[
P_\theta(\vb{y} | \vb{x}) = \sum_{\tilde{\vb{y}} \in B^{-1}(\vb{y})} P_{\theta}(\tilde{\vb{y}} | \vb{x})
\]
where $\tilde{\vb{y}} = {\tilde{y}}_1, \cdots, {\tilde{y}}_{T + L}$ is an alignment sequence,
$\tilde{y_i} \in \Sigma \cup \{\epsilon\}$, $T$ is the number of acoustic frames and $L$
is the number of labels. The function $B(\tilde{\vb{y}}) = \vb{y}$ removes the epsilons
from the alignment sequence. The alignment posterior is factorized along the alignment path as:
\[
P_{\theta}(\tilde{\vb{y}} | \vb{x}) = \prod_{j=1}^{T+L} P_{\theta}({\tilde{y}}_j | \vb{x}, {\tilde{y}}_{< j})
\]
and finally RNNT make the following assumption:
\[
P_{\theta}({\tilde{y}}_j | \vb{x}, {\tilde{y}}_{< j}) = P_{\theta} ({\tilde{y}}_j | \vb{x}, B({\tilde{y}}_{< j}) = y_{<u})
\]
which means if the prefix of two alignments be equal after epsilon removal, the model assigns same expansion probability
for the next alignment position. The inner terms in the above equation is defined  (Eq12-Eq15 of the RNNT paper):
\[
P_{\theta} ({\tilde{y}}_j | \vb{x}, B({\tilde{y}}_{< j}) = y_{<u}) =
\frac{\exp(W[{\tilde{y}}_j, :] \cdot (h_{j - u} + g_u) + b[{\tilde{y}}_j])}{\sum_{y \in \Sigma \cup \{ \epsilon \}} \exp(W[y, :] \cdot (h_{j - u} + g_u) + b[y])}
\]
where $h_{j-u}$ is the encoder activation at time frame $j-u$ (referred to as the transcription network in \cite{graves2012sequence}) and $g_u$ is the output of the prediction network
which is a simple stack of RNNs.

The equivalent GNAT model is configured as follow:
\begin{itemize}
    \item context dependency: $M$-gram where $M$ is the maximum value of $T + L$ in the training data set.
    \item alignment lattice: $k$-constrained label and frame dependent with $k$ being the label sequence length.
    \item weight function:
    \begin{itemize}
        \item $\omega_{\theta, \vb{x}}(q_a=(t, u), q_c=q, y={\tilde{y}}_{t+u+1}) \triangleq P_\theta ({\tilde{y}}_{t+u+1} | \vb{x}, B({\tilde{y}}_{< t+u +1}) = y_{< u})$
        \item locally normalized
        \item non-streaming
    \end{itemize}
\end{itemize}

While the original definition of the RNNT model is based on non-streaming encoder (transcription network), this model is widely used for streaming applications by using
a streaming encoder. This is in contradiction of the forward-backward derivations in the original paper which explicitly assumes dependency on the whole sequence for any position of alignment sequence (Eq17 of \cite{graves2012sequence})

Similar to RNNT, the hybrid autoregressive transducer (HAT) \cite{variani2020hybrid} model can be also configured in the GNAT framework with the same parametrization as RNNT. The only difference is the weight
function. The HAT model defines different probabilities for label transitions and
epsilon transitions (duration model in \cite{variani2020hybrid}):

\begin{eqnarray}
P_{\theta}({\tilde{y}}_j | \vb{x}, B({\tilde{y}}_{< j}) = y_{<u}) = 
  \begin{cases*}
    b_{j-u,u} & ${\tilde{y}}_{j} = \epsilon$ \\
      \left(1 - b_{j-u,u} \right) P_{\theta}\left(  {{y}}_{u + 1}  | X, B({\tilde{y}}_{< j}) = y_{<u} \right)        & ${\tilde{y}}_{j} \in \Sigma$
  \end{cases*}
  \label{eq:hat_model}
\end{eqnarray}
where $b_{t,u}$ is a sigmoid function defined in Eq6 of \cite{variani2020hybrid}.

\subsection{Supporting CTC Style Label Deduplication}
\label{sec:appendix_d}

The standard CTC model \cite{graves2006connectionist} is very similar to a GNAT model using a frame dependent alignment lattice and a $0$-gram context dependency.
One key difference is that CTC introduces a deduplication process when turning its model output sequence to a label sequence.
Each model output of a CTC model is either a lexical label from $\Sigma$, or the special $\epsilon$ (blank) label.
To obtain the label sequence, two steps are applied on the model output in order,
\begin{compactenum}
    \item Maximal consecutive repeated non-$\epsilon$ labels are merged into one (e.g. turning $a b b c$ into $a b c$, or $a b b \epsilon b$ into $a b \epsilon b$);
    \item All the $\epsilon$ labels are removed. 
\end{compactenum}

\begin{figure}
    \centering
    \scalebox{0.75}{%
    \begin{tikzpicture}
        \node[state, initial, accepting] (empty) {$\epsilon$};
        \node[state, accepting, below left of=empty, xshift=-1cm, yshift=-2cm] (a) {$a$};
        \node[state, accepting, below right of=empty, xshift=1cm, yshift=-2cm] (b) {$b$};

        \draw (empty) edge[loop above] node{$\epsilon:\epsilon$} (empty)
              (empty) edge[bend right, above left] node{$a:a$} (a)
              (empty) edge[bend left, above right] node{$b:b$} (b)
              (a) edge[right] node{$\epsilon:\epsilon$} (empty)
              (a) edge[loop left] node{$a:\epsilon$} (a)
              (a) edge[bend left, below] node{$b:b$} (b)
              (b) edge[left] node{$\epsilon:\epsilon$} (empty)
              (b) edge[bend left, above] node{$a:a$} (a)
              (b) edge[loop right] node{$b:\epsilon$} (b);
    \end{tikzpicture}
    }%
    \caption{An unweighted FST for CTC style label deduplication with $\Sigma=\{a,b\}$.}
    \label{fig:ctc-dedup-fst}
\end{figure}
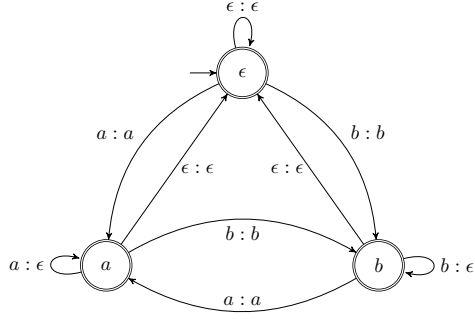

As a comparison, paths on the alignment lattice of the GNAT models in the main paper is equivalent to the model outputs in CTC, whereas the $\epsilon$ free label sequence seen by the context dependency is equivalent to the label sequence in CTC.
To support the deduplication of repeated non-$\epsilon$ labels, we need to introduce a finite state transducer into our series of finite state machine compositions.
Similar to a finite state automaton, a \emph{weighted finite-state transducer} (WFST) $T = (\Sigma, Q, i, F, \rho, E)$ over a semiring $\mathbb{K}$ is specified by a finite alphabet $\Sigma$, a finite set of states $Q$, an initial state $i \in Q$, a set of final states $F \subseteq Q$, a final state weight assignment $\rho : F \rightarrow \mathbb{K}$, and a finite set of transitions $E$.\footnote{Here we make the simplification that the input and output vocabularies are identical, i.e. $\Sigma$.}
The meaning of $\Sigma$, $Q$, $i$, $F$, and $\rho$ are identical to those of a WFSA.
The set of transitions $E$ is instead a subset of $Q \times (\Sigma \cup \{\epsilon\}) \times (\Sigma \cup \{\epsilon\}) \times \mathbb{K} \times Q$, i.e. containing a pair of input/output labels instead of just one.
A WFSA can be viewed as a WFST with identical input/output labels on each arc, and similar to WFSA intersection, a series of WFST can be composed into a single WFST.
We refer the readers to \cite{mohri2009} for a full description of WFST and the composition algorithm.
Figure $\ref{fig:ctc-dedup-fst}$ is an example unweighted FST, when composed with another input FSA or FST, performs the CTC style label deduplication.
More generally, the unweighted label deduplication transducer $D$ of vocabulary $\Sigma$ consists of
\begin{compactitem}
    \item States $Q_D = \Sigma \cup \{ \epsilon \}$, $i_D = \epsilon$, $F_D = Q_D$
    \item Transitions $E = \{ (p, x, x, x) | p, x \in Q_D \} \cup \{ (x, x, \epsilon, x) | x \in \Sigma \}$
\end{compactitem}

Given a context dependency FSA $C$, an alignment lattice FSA $L_T$, and the weight function $\omega_{\theta,\vb{x}}$, as defined in Section \ref{subsec:inducing-the-wfsa}, a GNAT model with CTC style label deduplication induces a WFST $T_{\theta,\vb{x}}$ as follows,

\begin{align*}
    Q_{\theta,\vb{x}} & = Q_T \times Q_D \times Q_C \\
    i_{\theta,\vb{x}} & = (i_T, i_D, i_C) \\
    F_{\theta,\vb{x}} & = F_T \times F_D \times F_C \\
    \begin{split}
        E_{T_{\theta,\vb{x}}}  = & \Bigl\{\bigl((q_a, q_d, q_c), y, y, \omega_{\theta,\vb{x}}(q_a, q_c, y), (q_a', q_d', q_c')\bigr) \mid \\ & y \in \Sigma,\ (q_a, y, q'_a) \in E_T,\ (q_d, y, y, q_d') \in E_D,\ (q_c, y, q'_c) \in E_C\Bigr\} \\
            & \cup \Bigl\{\bigl((q_a, q_d, q_c), \epsilon, \epsilon, \omega_{\theta,\vb{x}}(q_a, q_c, \epsilon), (q_a', q_d', q_c)\bigr) \mid \\
            & (q_a, \epsilon, q'_a) \in E_T,\ (q_d, \epsilon, \epsilon, q_d') \in E_D,\ q_c \in Q_C\Bigr\} \\
            & \cup \Bigl\{ \bigl((q_a, q_d, q_c), y, \epsilon, \omega_{\theta,\vb{x}}(q_a, q_c, y), (q_a', q_d', q_c)\bigr) \mid \\ & y \in \Sigma,\ (q_a, y, q_a') \in E_T,\ (q_d, y, \epsilon, q_d') \in E_D \Bigr\}
    \end{split}  \\
    \rho_{T_{\theta,\vb{x}}}(q) & = \bar{1},\ \forall q \in F_{\theta,\vb{x}}
\end{align*}

In other words, the topology of $T_{\theta,\vb{x}}$ is the same as the following cascade of FST compositions,
\begin{compactenum}
    \item $D \cdot C$ treating output $\epsilon$ labels in $D$ as empty (i.e. standard FST composition).
    \item $L_T \cdot (D \cdot C)$ treating $\epsilon$ transitions in $L_T$ and input $\epsilon$ labels in $(D \cdot C)$ as regular labels. 
\end{compactenum}
and the transition weights are defined using $\omega_{\theta,\vb{x}}$ just like the GNAT models in the main paper.

When implemented naively, CTC style label deduplication causes a $|Q_D| = (V+1)$ blow up in $|Q_{\theta,\vb{x}}|$.
However, by inferring about states in $Q_D$ from states in $Q_C$, we can greatly reduce the number of states needed.
For each state $x \in \Sigma$ in $Q_D$, we know the last non-$\epsilon$ label observed when reaching state $x$ must be label $x$.
Similarly, for context dependencies we care about ($n$-gram and string), there is a unique label $x(q_c)$ for all incoming arcs of each non-start state $q_c$ (start states do not have any incoming arcs in these context dependencies).
Thus, the states in $Q_{\theta,\vb{x}}$ that are reachable from the start must match one of the following patterns,
\begin{compactitem}
    \item $(q_a, \epsilon, q_c),\ \forall q_a \in Q_A, q_c \in Q_X$
    \item $(q_a, x(q_c), q_c),\ \forall q_a \in Q_A, q_c \in Q_X \setminus \{i_X\}$ 
\end{compactitem}
This means the actual number of states we shall visit in computing the shortest distance is only $2|Q_A||Q_C|$.
\section{Accelerator-Friendly Computation}
\label{sec:accelerator_friendly_modeling}

The standard shortest distance/path algorithm for acyclic WFSA \cite{mohri2002} can be used for training (computing $W(A)$ for some acyclic $A$) and inference of a GNAT model.
To compute $W(A)$ for an acyclic WFSA $A$, we maintain the following forward weight $\alpha_q$ for each state $q$ in $Q_A$:
\[
    \alpha_q = \left\{
    \begin{array}{ll}
        \bar{1} & \text{if } q = i_A, \\
        \bigoplus_{(p, y, w, q) \in E_A} \alpha_p \otimes w & \text{else}.
    \end{array} \right.
\]
The weight of $A$ is then $W(A) = \bigoplus_{q \in F_A} \alpha_q \otimes \rho_A(q)$.
The recurrence in the definition of $\alpha_q$ can be computed by visiting states in $Q_A$ in a topological order.

To make better use of the compute power of modern accelerator hardware, we observe the following properties of the $C$ or $L_T$ presented so far that enable us to use a more vectorized variant of the shortest distance algorithm in Figure~\ref{fig:vectorized-shortest-distance}:
\begin{compactitem}
    \item From any topological ordering on $Q_T$, we can derive a topological ordering on $Q_{\theta,\vb{x}}$.
    \item The $n$-gram context dependency FSA $C_n$ is deterministic, namely leaving any state there is no more than 1 transition for any label $y \in \Sigma$, and there is no $\epsilon$-transition.
    \item For all three types of alignment lattices, for any non-final state $q \in Q_T \setminus F_T$, there is a unique next state $\operatorname{succ}(q)$ for transitions leaving $q$ consuming any label $y \in \Sigma$.
\end{compactitem}

\begin{figure}
    \centering
    \begin{algorithmic}
        \STATE\COMMENT{Initialize the length $|Q_C|$ forward weight vectors $\bar{\alpha}_{q_a}$}
        \FORALL{$q_a \in Q_T$}
            \STATE{$\bar{\alpha}_{q_a} \gets [\bar{0}, \dots, \bar{0}]$}
        \ENDFOR
        \STATE{$\bar{\alpha}_{i_T}[i_C] \gets \bar{1}$}
        \STATE\COMMENT{Compute $\bar{\alpha}_{q_a}$ for $q_a \neq i_T$}
        \FORALL{$q_a \in Q_T$ in topological order}
            \STATE\COMMENT{$\Omega$ is a $[|Q_C|, |\Sigma|+1]$ matrix}
            \STATE{$\Omega \gets \bar{\omega}_{\theta,\vb{x}}\bigl(q_a, Q_C, \Sigma \cup \{ \epsilon \}\bigr)$}
            \IF{$q_a$ has outgoing label transitions to $q_a' = \operatorname{succ}(q_a)$}
                \STATE{$\bar{\alpha}_{q_a'} \gets \bar{\alpha}_{q_a'} \bar{\oplus} \operatorname{next}_C\bigl(\bar{\alpha}_{q_a}, \Omega[:, \Sigma]\bigr)$}
            \ENDIF
            \FORALL{$q_a'$ such that $(q_a, \epsilon, q_a') \in E_T$}
                \STATE{$\bar{\alpha}_{q_a'} \gets \bar{\alpha}_{q_a'} \bar{\oplus} \bigl(\bar{\alpha}_{q_a} \bar{\otimes} \Omega[:,\epsilon]\bigr)$}
            \ENDFOR
        \ENDFOR
        \RETURN{$\bigoplus_{q_a \in F_T, q_c \in F_C} \bar{\alpha}_{q_a}[q_c]$}
    \end{algorithmic}
    \caption{The vectorized shortest distance algorithm for $A_{\theta,\vb{x}}$. We denote $\bar{\oplus}$, $\bar{\otimes}$, and $\bar{\omega}_{\theta,\vb{x}}$ the vectorized versions of the corresponding operations.}
    \label{fig:vectorized-shortest-distance}
\end{figure}

Center to an efficient implementation of the algorithm in Figure~\ref{fig:vectorized-shortest-distance} is the function $\operatorname{next}_C$.
This function receives as input the current forward weight vector $\bar{\alpha}_{q_a}$ for states $(q_a, q_c),\ \forall q_c \in Q_C$, and the transition weights for leaving these states via label transitions, and returns the forward weights going to states $(q_a', q_c')$ by taking the $(q_c, y, q_c')$ transitions for $y \in \Sigma$.
In other words, $\operatorname{next}_C[q_c'] = \bigoplus_{(q_c, y, q_c') \in E_C} \bar{\alpha}_{q_a}[q_c] \otimes \Omega[q_c, y]$.
The $n$-gram context dependency $C_n$ allows a particularly simple and efficient implementation of $\operatorname{next}_{C_n}$, as outlined in Figure~\ref{fig:n-gram-next}.
The key observation is that when we number the states in $Q_C$ following the lexicographic order, the $|\Sigma|$ transitions leaving the same $q_c$ lead to states in a consecutive range $[\sigma(q_c y_0), \dots, \sigma(q_c y_{|\Sigma|-1})]$, where $[y_0, \dots, y_{|\Sigma|-1}]$ are the lexicographically sorted labels of $\Sigma$, and $\sigma(s)$ is the suffix of label sequence $s$ of length up to $n-1$.

\begin{figure}
    \centering
    \begin{algorithmic}
        \STATE\COMMENT{Inputs: $\bar{\alpha}_{q_a}$ and $\Omega[:, \Sigma]$}
        \IF{$n = 1$}
            \STATE\COMMENT{$Q_{C_n}$ contains only $i_{C_n}$}
            \RETURN{$\bar{\alpha}_{q_a} \bar{\otimes} \bigoplus_{y \in \Sigma} \Omega[i_{C_n}, y]$}
        \ENDIF
        \STATE\COMMENT{Initialize length $|Q_{C_n}|$ vector $\bar{\alpha}$}
        \STATE{$\bar{\alpha} \gets [\bar{0}, \dots, \bar{0}]$}
        \STATE\COMMENT{States in $Q_{C_n}$ are numbered from $0$ to $|Q_{C_n}|-1 = \sum_{i=1}^{n-1} |\Sigma|^i$ following the lexicographic order}
        \STATE{$l \gets 0$}
        \FOR{$i = 0$ \TO $n-2$}
            \STATE{$h \gets l + |\Sigma|^i$}
            \STATE{$\bar{\alpha}[l\cdot|\Sigma|+1 : h\cdot|\Sigma|+1] \gets \operatorname{flatten}(\Omega[l:h, \Sigma])$}
            \STATE{$l \gets h$}
        \ENDFOR
        \FOR{$i = 0$ \TO $|\Sigma| - 1$}
            \STATE{$\bar{\alpha}[l:] \gets \bar{\alpha}[l:] \bar{\oplus} \operatorname{flatten}(\Omega[l+i\cdot|\Sigma|^{n-2}:l+(i+1)\cdot|\Sigma|^{n-2}, \Sigma])$}
        \ENDFOR
        \RETURN{$\bar{\alpha}$}
    \end{algorithmic}
    \caption{Specialized implementation of $\operatorname{next}_{C_n}$. The $\operatorname{flatten}$ function flattens a matrix into a vector by joining the rows.}
    \label{fig:n-gram-next}
\end{figure}

During training, we also need to compute the shortest distance $D(A_{\theta,\vb{x}} \cap \vb{y})$.
We note the algorithm in Figure~\ref{fig:vectorized-shortest-distance} can also be used for this purpose since $(L_T \cap C) \cap \vb{y} = L_T \cap (C \cap \vb{y})$, and we simply need to substitute $C$ with $C \cap \vb{y}$ in the algorithm.

\section{Memory and Computation Time Benchmarks}
\label{sec:appendix_c}

The memory and computation benchmark of our implementation for the GNAT model is presented
in Table~\ref{tab:benchmarks}. We present benchmarks for training and inference for different
configurations of the GNAT model:
\begin{itemize}
    \item Context dependency: $0$-gram, $1$-gram and $2$-gram
    \item Alignment lattice: frame dependent, 1-constrained label and frame dependent
    \item Weight functions: Per-state linear projection (unshared), Shared linear projection with per-state embedding (shared-emb), Shared linear projection with RNN state embedding (shared-rnn)
\end{itemize}

\begin{table*}[hbt!]
\begin{center}
\begin{tabular}{|c|c|c|c|c|c|c|c|}
\hline
context    & alignment  & \multicolumn{2}{c|}{weight function} & \multicolumn{2}{c|}{memory [M]} & \multicolumn{2}{c|}{time [sec]}\\
\cline{3-8}
dependency &  lattice   &    type   & normalization    & train & decode & train & decode \\
\hline
\multirow{8}{*}{0-gram} & \multirow{6}{*}{frame} & \multirow{2}{*}{unshared}         & local & 126.47 & 64.97 & 0.15 & 0.02 \\
\cline{4-8}
                   &                        &                                   & global & 124.58 & 65.19 & 0.14 & 0.02\\
\cline{3-8}
                   &                        & \multirow{2}{*}{shared-emb}       & local & 124.20 & 65.12 & 0.20 & 0.02\\
\cline{4-8}
                   &                        &                                   & global & 124.64 & 65.19 & 0.18 & 0.02\\
\cline{3-8}
                   &                        & \multirow{2}{*}{shared-rnn}       & local  & 124.43 & 65.16 & 0.20 & 0.02\\
\cline{4-8}
                   &                        &                                   & global & 124.88 & 65.29 & 0.18 & 0.02\\
\cline{2-8}
                   & \multirow{6}{*}{label frame} & \multirow{2}{*}{unshared}   & local & 174.62 & 65.00 & 0.16 & 0.02\\
\cline{4-8}
                   &                              &                             & global & 172.21 & 65.20 & 0.18 & 0.04 \\
\cline{3-8}
                   &                              & \multirow{2}{*}{shared-emb} & local & 172.32 & 65.15 & 0.20 & 0.02\\
\cline{4-8}
                   &                              &                             & global & 172.25 & 65.17 & 0.22 & 0.03\\
\cline{3-8}
                   &                              & \multirow{2}{*}{shared-rnn} & local & 172.55 & 65.19 & 0.20 & 0.02\\
\cline{4-8}
                   &                              &                             & global & 172.49 & 65.27 & 0.22 & 0.04\\
\hline
\multirow{8}{*}{1-gram} & \multirow{6}{*}{frame} & \multirow{2}{*}{unshared}         & local  & 144.04  & 64.95 & 0.17 & 0.044\\
\cline{4-8}
                   &                        &                                   & global & 146.23 & 65.51  & 0.19 & 0.05\\
\cline{3-8}
                   &                        & \multirow{2}{*}{shared-emb}       & local  & 156.02 & 70.42 & 0.22 & 0.05 \\
\cline{4-8}
                   &                        &                                   & global & 158.12 & 70.52 & 0.22 & 0.05\\
\cline{3-8}
                   &                        & \multirow{2}{*}{shared-rnn}       & local  & 157.04 & 70.67 & 0.23 & 0.05\\
\cline{4-8}
                   &                        &                                   & global & 159.15 & 70.76 & 0.23 & 0.05\\
\cline{2-8}
                   & \multirow{6}{*}{label frame} & \multirow{2}{*}{unshared}   & local  & 192.19 & 64.98 & 0.18 & 0.04\\
\cline{4-8}
                   &                              &                             & global & 192.70  & 65.19 & 0.27 & 0.07\\
\cline{3-8}
                   &                              & \multirow{2}{*}{shared-emb} & local & 204.13 & 70.45 & 0.23 & 0.05\\
\cline{4-8}
                   &                              &                             & global & 204.65 & 70.14 & 0.29 & 0.07\\
\cline{3-8}
                   &                              & \multirow{2}{*}{shared-rnn} & local & 205.16  & 70.69 & 0.23 & 0.05\\
\cline{4-8}
                   &                              &                             & global & 205.68 & 70.38 & 0.30 & 0.07\\
\hline
\multirow{8}{*}{2-gram} & \multirow{6}{*}{frame} & \multirow{2}{*}{unshared}         & local & 306.94 & 187.21  & 0.23 & 0.07\\
\cline{4-8}
                   &                        &                                   & global & 513.58 & 195.36 & 1.55 & 0.41\\
\cline{3-8}
                   &                        & \multirow{2}{*}{shared-emb}       & local & 156.05 & 70.42 & 0.23 & 0.05\\
\cline{4-8}
                   &                        &                                   & global & 181.42 & 73.49 & 1.16 & 0.23\\
\cline{3-8}
                   &                        & \multirow{2}{*}{shared-rnn}       & local  & 174.22 & 73.04 & 0.23 & 0.05\\
\cline{4-8}
                   &                        &                                   & global & 199.62 & 76.11 & 1.16 & 0.23\\
\cline{2-8}
                   & \multirow{6}{*}{label frame} & \multirow{2}{*}{unshared}   & local & 320.98 & 187.24 & 0.24 & 0.07\\
\cline{4-8}
                   &                              &                             & global & 428.40 & 187.90 & 3.79 & 0.94\\
\cline{3-8}
                   &                              & \multirow{2}{*}{shared-emb} & local & 204.17 & 70.45 & 0.24 & 0.05\\
\cline{4-8}
                   &                              &                             & global & 210.61 & 71.42 & 2.63 & 0.48\\
\cline{3-8}
                   &                              & \multirow{2}{*}{shared-rnn} & local & 222.33 & 73.07 & 0.24 & 0.05\\
\cline{4-8}
                   &                              &                             & global & 229.17 & 74.04 & 2.64 & 0.49\\
\hline
\end{tabular}
\end{center}
\caption{\label{tab:benchmarks} Memory and computation benchmarks of the GNAT model for
different configurations.}
\end{table*}

For each configuration, the memory usage footprint is presented
in terms of MB and total computation time is presented in terms
of number of seconds. The benchmarks do not include the memory and computation footprint of the encoder activations. The training benchmarks are corresponding to the calculation of the GNAT criterion as well as all the backward gradient calculation up to the encoder activations. The evaluation benchmarks only contain the forward pass memory and compute footprint to find the most likely hypothesis.

All the memory and computation benchmarks are evaluated for an input batch of $32$ examples each with $1024$ number of frames.
Each frame is a $512$-dim vector corresponding to the encoder activations. Each example in the input batch are assumed to have at most $256$ labels. The alphabet size is set to $32$.

The main observations are:
\begin{itemize}
    \item The larger context dependency lead to more memory and compute footprint. This is expected since the computation complexity is directly
    related to the context dependency state size. However, interestingly, the memory and computation values do not scale exponentially by value of $n$
    in $n$-gram context dependency (as a result by number of states in the context dependency).
    \item label frame dependent alignment lattice generally leads to higher memory usage and computation time compare to the frame dependent alignment
    lattice. This is expected since the label frame dependent consist of alignment paths of length $1024 + 256 = 1270$, corresponding to the sum of
    number of frames and number of labels.
    \item The per-state linear projection weight function requires more memory and has longer compute time compare to the shared weights function
    which is expected by design. Both shared weight functions are performing on-par of each other in terms of memory and compute.
    \item The global normalization requires more memory and time and the difference is more significant for context dependency FSAs with more number of states ($2$-gram versus $1$-gram).
\end{itemize}


\end{document}